\documentclass{article}

\usepackage[preprint]{FMC_Paper}

\usepackage[utf8]{inputenc} 
\usepackage[T1]{fontenc}    
\usepackage{hyperref}       
\usepackage{url}            
\usepackage{booktabs}       
\usepackage{amsfonts}       
\usepackage{nicefrac}       
\usepackage{microtype}      

\usepackage{graphicx}
\usepackage{caption}
\usepackage{natbib}

\title{Solving Atari Games Using Fractals And Entropy}

\author{
  Sergio Hernandez Cerezo \\
  HCSoft Programación, S.L., 30007 Murcia, Spain\\
  \texttt{sergio@hcsoft.net} \\
  \And
  Guillem Duran Ballester \\
  \texttt{guillem.db@gmail.com} \\
  \And
  Spiros Baxevanakis \\
  \texttt{spiros.baxevanakis@gmail.com}
}

\begin{document}

\maketitle

\begin{abstract}
  In this paper we introduce a novel MCTS based approach that is derived from the laws of the thermodynamics. The algorithm, coined Fractal Monte Carlo (FMC), allows us to create an agent that takes intelligent actions in both continuous and discrete environments while providing control over every aspect of the agent's behavior. Results show that FMC is several orders of magnitude more efficient than similar techniques, such as MCTS, in the Atari games tested.
\end{abstract}

\section{Introduction}
Artificial intelligence methods are currently limited by the lack of a concrete definition of intelligence that would assist in creating agents that exhibit intelligent behavior. Fractal AI theory (FAI) \citep{cerezo2018fractal} is inspired by the work of \citeauthor{wissner2013causal} who proposed the concept of Causal Entropic Forces and showed that an agent exhibits intelligent behavior when it tries to maximize its Causal Path Entropy or equivalently, maximize its future freedom of action. In order to achieve that, the agent directly modifies its degrees of freedom in such a way that it assumes the state with the highest number of possible, different futures. To find the available futures, the agent will need to scan the action space and recreate the Causal Cone that contains the paths of all possible future internal configurations that start from its initial state. This scanning process is called a Scanning Policy. Available actions are then assigned a probability of being chosen by the Deciding Policy.
In Fractal AI, intelligence is defined as the ability to minimize a sub-optimallity coefficient based on the similitude of two probability distributions created from the scanning and decision policies.

Using the principles described in FAI theory we developed a Monte Carlo approach coined Fractal Monte Carlo (FMC) that is based on the second law of thermodynamics. The algorithm develops a swarm of walkers that evolve in an environment while balancing exploitation and exploration by means of a mechanism named cloning. This process generates a "fractal tree" that will tend to fill up all the causal cone, from the interconnected paths of the walkers. The algorithm can be applied to both continuous and discrete decision spaces while remaining extremely efficient.

In addition, FMC provides a suite of parameters that control computational resources as well as agent reaction time. To test our algorithm we put it up against 55 different Atari environments and compared our results to state of the art algorithms like A3C \citep{mnih_asynchronous_2016}, NoisyNet \citep{fortunato_noisy_2017}, DQN \citep{mnih2013playing,mnih_human-level_2015} and variants.
\\

\section{Related work}
Recently there have been numerous breakthroughs in reinforcement learning most of them originating from Deepmind. They created an end to end model free reinforcement learning technique, named Deep Q Learning \citep{mnih2013playing,mnih_human-level_2015}, that scored astonishingly well and outperformed previous approaches in Atari games. Their Deep Q Learner achieves such feats by estimating Q-values directly from images and stabilizes learning by means such as experience replay and frame skipping. Later in 2016 they created AlphaGo which went to beat the world champion in the game of Go \citep{silver2016mastering} and AlphaChem which was shown to outperform hardcoded heuristics used in retrosynthesis \citep{segler2017towards}. Both AlphaGo and AlphaChem use some form of deep reinforcement learning in conjunction with a MCTS variant, UCP \citep{isasi2014schemata}.
\\

\section{Background}
FMC is a robust path-search algorithm that efficiently approximates path integrals formulated as a Markov decision process by exploiting the deep link between intelligence and entropy maximization \citep{wissner2013causal} that naturally produces an optimal decision-making process. FMC formulates agents that exhibit intelligent behavior in Atari game emulators. Such agents create a swarm of walkers that explores the Causal Cone and eventually, when the time horizon is met, select an action based on the walkers' distribution over the action space.

\subsection{Causal cones}
In order to find the best path, Fractal Monte Carlo scans the space of possible future states thereby constructing a tree which consists of potential trajectories that describe the future evolution of the system. We define a Causal Cone $X(x_0, \tau)$ as the set of all possible paths the system can take starting from an initial state $x_0$ if allowed to evolve over a time interval of length $\tau$, the 'time horizon' of the cone. A Causal Cone can be divided into a set of Causal Slices defined as $X_H(x_0, t)$, where each Causal Slice contains all the possible future states of the paths at a given time $t$. If $t=\tau$ the Causal Slice is called the cone’s ‘horizon’ and contains the final states. The rest of the cone, where $t<\tau$, is usually referred to as the cone’s ‘bulk’.
\\

\begin{figure}[h]
  \centering
  \includegraphics[width=\textwidth]{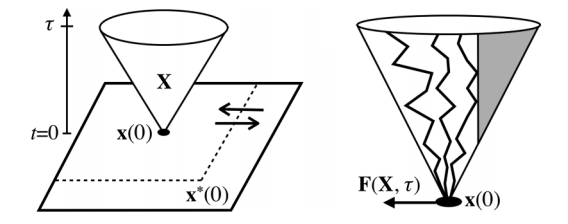}
  \caption{(from \cite{wissner2013causal}): A Causal Cone visualization. On the left, the Causal Cone expands from the initial $x(0)$ point and all the possible future paths are expanded upwards in time. On the right the entropic force compels the walkers to avoid the grey excluded volume and therefore evolve on the remaining space.}
  \label{causal-cone}
\end{figure}

\subsection{Dead and alive statuses}
The death condition is a flag set by the programmer that lets us incorporate arbitrary boundary conditions to the behaviour of the agent and helps the swarm avoid undesired regions of the state space. We will assume an external death condition is defined over $E$ so a portion of the space can be forbidden for the system, as shown in Figure \ref{causal-cone} . We will consider a state inside this excluded region "dead" while all other states are "alive".

\subsection{Reward function}
Agents make decisions based on a non-negative reward function that (we assume) is defined over the state space. For every slice $X_H(x_0,t)$ of the causal cone, we can calculate the total reward $R_{TOT}(x_0,t)$ of the slice as the integral of the reward over the slice. We may then convert the reward into a probability density $P_R$ over the slice as follows:
\begin{equation}
  P_R(x\mid x_0,t) = R(x)/R_{TOT}(x_0,t)
\end{equation}
The general idea behind the algorithm will be that, the density distribution of the scanning should match the reward density distribution of the state space.

\subsection{Policies}
The proposed algorithm uses a set of two policies to choose and score actions. First we define a scanning policy $\pi_s$ that, given a swarm of initially identical states, defines its possible evolution over time as a stochastic process.
\\
After the scanning is finished we need a deciding policy $\pi_D$ that will assign a probability of being chosen to each action:
\begin{equation}
  \pi_D : A \rightarrow [0,1]
\end{equation}
In order to measure how different two probability distributions are we will use a modified version of the Kullback-Leibler divergence:
\begin{equation}
  D_H(P\mid\mid Q) = log(\Pi(2 - p_i^{p_i})/\Pi(2 - q_i^{p_i}))
\end{equation}
This divergence is well defined for any possible distributions $p$ and $q$, including the problematic case when $(p_i > 0, q_i = 0). $

\section{Fractal Monte Carlo}
Fractal Monte Carlo is a path-search algorithm derived from Fractal AI theory \citep{cerezo2018fractal} that produces intelligent behavior by maximizing an intrinsic reward represented as Causal Path Entropy \citep{wissner2013causal}. When making a decision, Fractal Monte Carlo (FMC) establishes a tree that describes the future evolution of the system. This tree is expanded by a swarm of walkers that populates its leaf nodes.  The swarm will undergo an interactive cellural automaton like process in order to make the tree grow efficiently. When a maximum amount of computation has been reached, the utility of each action will be considered proportional to the number of walkers that populate leaf nodes originating from the same action.
\\

\subsection{The algorithm}
FMC steps are outlined below:

STEP 1: Initialize the walkers to the root state.\\

STEP 2: \emph{Perturb} the swarm of walkers.\\

STEP 3: \emph{Evaluate }the position of each walker with respect to the whole swarm.\\

STEP 4: \emph{Recycle} the walkers that are in a dead state or have been poorly valued against the rest of the swarm.\\

STEP 5: Repeat phases 2-4 until we reach maximum computational resources.\\

STEP 6: \emph{Choose} the action with the highest utility.\\
\\
In more detail:

Perturb:
\begin{enumerate}
  \item Every walker chooses a random action and acts in the environment.
\end{enumerate}

Evaluate:
\begin{enumerate}
  \item For every walker $A$ select an alive walker $B$ at random and measure the euclidean distance $D_i$ between their observations.
  \label{step1}
  \item Normalize distances and rewards using the "Relativize" function
  \item Calculate the virtual reward of each walker. We define virtual reward $VR$ at a state $W_i$ as:
  \begin{equation}
    VR_i=R_i*D_i
  \end{equation}
  Where $R_i$ is the reward value at state $W_i$.
  Virtual reward is a stochastic measure of the importance of a given walker in respect to the whole swarm.
\end{enumerate}

Recycle:
\begin{enumerate}
  \item Each walker A is compared to another randomly selected walker C and gets assigned a probability of cloning to the leaf node of walker C.
  \item Determine if the walker A will clone to C based on the cloning probability and the death condition.
  \item Finally transfer the walkers that are set for cloning to their target leaf node.
\end{enumerate}

Choose:
\begin{enumerate}
  \item When assigning a utility value to an option, FMC counts how many walkers took each option at the root state. To choose an action in the continuous or general case we calculate the average of the actions weighted by their normalized utilities or scores. In the discrete case, the action that approximates better the aforementioned average is chosen.
\end{enumerate}

Relativize :
\begin{enumerate}
  \item Given $R$ real values we calculate the mean $R_m$ and the standard deviation $R_d$.
  \item Normalize the values using:
  \begin{equation}
    R_i' = (R_i - R_m) / R_d
  \end{equation}
  \item Reshape the values into a Gaussian N(0,1) distribution.
  \item Then scale using: if $R_i' < 0$ then $R_i' = \exp(R_i')$ else $R_i' = 1 + ln(R_i')$

\end{enumerate}

\subsection{Parameters}

\textbf{Time Horizon} sets a limit for how far in the future the walkers of the Swarm will foresee the aftermath of their initial actions. In other words, the walkers will seek to meet their set Time Horizon when going deeper in the tree but never go past it. The ideal Time Horizon value allows an agent to see far enough in the future to detect which actions lead inevitably to death.

\textbf{Max samples} is an upper bound on computational resources. It limits the number of times that FMC can make a perturbation to build a causal cone. The algorithm will try to keep computational resource usage as low as it can providing it meets the time horizon. A good guide to setting this parameter is $Number\_of\_walkers * time\_horizon * Repeat Actions$, with $Repeat Actions=5$ a number that works well in Atari games but highly depends on the task.

\textbf{Number of walkers} represents the maximum number of paths that FMC will simulate. This number is related to "how thick" we want the resulting representation of the causal cone to be. The algorithm will try to use the maximum number of walkers possible.

\textbf{Time step (dt)} is the time interval the agents keep each decision made. Time horizon / dt will define the number of steps to be taken by walkers.

\subsection{Time complexity}
Computational time complexity of the algorithm can be shown to be of $O(n)$ with $n = number\_of\_steps * number\_of\_walkers$.
\\

\section{Findings}
In this section we present our performance results in Atari games and then compare our approach to MCTS and other state of the art learnin-based approaches in eight games.

\subsection{Atari environments}
We tested our algorithm in 55 different Atari games using the OpenAI gym toolkit. In most games the agent used RAM data as observations to make decisions. As seen in table \ref{table1}, our results show that Fractal Monte Carlo outperforms previous state of the art (SOtA) approaches in 49 of those (89\%). In each game we choose the appropriate parameters by experimentation and intuition. The extensive table containing the exact parameters used for every game is available in our github repository \citep{fractalai_repo}. Furthermore, we compared FMC performance when using RAM data versus using images as observations in eight games and found that there is an overall $61\%$  performance difference in favor of RAM data(Table \ref{table2}).

\begin{table}[h]
  \centering
  \caption{Fractal Monte Carlo in Atari games}
  \label{table1}
  \begin{tabular}{|l|l|l|}
    \hline
    Available games              & 55 &          \\
    \hline
    Games played by FMC          & 55 & 100.00\% \\
    \hline
    FMC better than avg human    & 51 & 92.73\%  \\
    \hline
    FMC better than SOtA         & 49 & 89.09\%  \\
    \hline
    Solved or above human record & 25 & 45.45\%  \\
    \hline
    Solved due to the 1M bug     & 11  & 20.00\%  \\
    \hline
  \end{tabular}
  \caption*{In  51 out of 55 games FMC scored better than the average human meaning a human that has played for 2 hours. FMC solved or scored higher than the human record in $45\%$ of the games tested. A game is considered 'solved' when we hit an eventual score limit or we can play it endlessly. SOtA is an abbreviation for "State Of the Art". The term "1M bug" refers to some games having a hardcoded score limit, usually found at 999,999.}
\end{table}

\begin{table}[h]
  \centering
  \caption{Image data versus RAM dump}
  \label{table2}
  \begin{tabular}{|l|l|l|l|l|}
    \hline
    Environment   & IMG      & RAM        & RAM vs IMG  & RAM vs SoTA \\
    \hline
    atlantis      & 145000   & 139500     & 96.21\%     & 1.59\%      \\
    \hline
    bank heist    & 160      & 280        & 175.00\%    & 0.68\%      \\
    \hline
    boxing        & 100      & 100        & 100.00\%    & 100.60\%    \\
    \hline
    centipede     & immortal & immortal   & 100.00\%    & 2373.87\%   \\
    \hline
    ice hockey    & 16       & 33         & 206.25\%    & 311.32\%    \\
    \hline
    ms pacman     & 23980    & 29410      & 122.64\%    & 468.09\%    \\
    \hline
    qbert         & 17950    & 22500      & 125.35\%    & 358.11\%    \\
    \hline
    video pinball & 273011   & 999999     & 366.29\%    & 105.31\%    \\
    \hline
                  &          &            & 161.47\%    & 464.95\%    \\
    \hline
  \end{tabular}
  \caption*{Comparison of FMC scores when using IMG and RAM data as observations in eight Atari games. Parameters used are: Repeat Actions=5, Time Horizon=15, Max Samples=300, Number of Walkers=30. We find that, in total, RAM yields $61.47\%$ better results than IMG scores and outperforms State Of the Art (SOtA) methods in most games tested.}
\end{table}

\subsection{Comparison against UCT}
In table \ref{table3} we compare FMC with the state of the art MCTS implementation UCT on the only eight games we found to be solved in the literature \citep{guo2014deep}. The results presented in table 3 show that our method our method clearly outperforms MCTS while being three to four orders of magnitude more efficient.

\begin{table}[h]
  \centering
  \caption{Comparison against MCTS variant, UCT}
  \label{table3}
  \begin{tabular}{|l|l|l|l|l|l|l|}
    \hline
    Game           & \multicolumn{3}{c|}{Scores}  & \multicolumn{3}{c|}{Samples per step} \\
    \hline
    & MCTS    & FMC     & \%      & MCTS        & FMC     & Efficiency   \\
    \hline
    Asterix        & 226,000 & 999,500 & 442\%   & -----       & 241     & -----        \\
    \hline
    Beam rider     & 7,233   & 288,666 & 3991\%  & 3,000,000   & 946     & x 3,171      \\
    \hline
    Breakout       & 406     & 864     & 213\%   & 3,000,000   & 866     & x 3,386      \\
    \hline
    Enduro         & 788     & 5,279   & 670\%   & 4,000,000   & 390     & x 10,256     \\
    \hline
    Pong           & 21      & 21      & 100\%   & 150,000     & 158     & x 949        \\
    \hline
    Q-bert         & 18,850  & 999,999  & 3523\%   & 3,000,000   & 3,691   & x 813      \\
    \hline
    Seaquest       & 3,257   & 999,999 & 30703\% & 3,000,000   & 964     & x 3,112      \\
    \hline
    Space invaders & 2,354   & 17,970   & 763\%   & 3,000,000   & 1,830     & x 1,639      \\
    \hline
  \end{tabular}
  \caption*{Comparison of MCTS versus FMC performance in eight Atari Games. We also compare the number of simulations or "samples" each algorithm used per step. The metric "efficiency" is calculated as MCTS samples per step / FMC samples per step for each game}
\end{table}

\section{Conclusions}
In this work, we propose a new thinking framework called Fractal AI theory that we used to define intelligent behavior. FAI principles provided a basis for creating a new Monte Carlo approach based on maximizing Causal Path Entropy. We put up our algorithm against Atari environments and our results showed that it performs better than state of the art algorithms, like DQN and its variants, in most games.

Our algorithm has many potential applications especially for improving methods that use MCTS. Furthermore FMC can produce high amount and high quality training data for use in training reinforcement learning agents. For example, deep Q agents learn to associate reward expectations with states after being trained on a huge amount of data mainly consisting of random rollouts. FMC high quality rollouts can be fed into a DQN in the training stage which might result in a boost in training performance. Another promising idea for improving FMC is to add learning capabilities to walkers using a neural network. The network would be trained on correct decisions made by the agent and output a probability distribution over the action space. Then walkers would sample that distribution instead of picking a random action thus transitioning into an informed search model.

Fractal Monte Carlo is a just one of the possible algorithms inspired by FAI theory. Much research is still needed to explore other possible implementations of this new concept and their many potential applications in the real world.

\nocite{*}
\bibliography{FMC_Paper}

\begin{thebibliography}{}

\bibitem[\protect\astroncite{Bellemare et~al.}{2016}]{bellemare2016unifying}
Bellemare, M., S.~Srinivasan, G.~Ostrovski, T.~Schaul, D.~Saxton, and
  R.~Munos\leavevmode\nopagebreak\newline 2016.
\newblock Unifying count-based exploration and intrinsic motivation.
\newblock In {\em Advances in Neural Information Processing Systems}, Pp.~
  1471--1479.

\bibitem[\protect\astroncite{Bellemare
  et~al.}{2017}]{bellemare2017distributional}
Bellemare, M.~G., W.~Dabney, and R.~Munos\leavevmode\nopagebreak\newline 2017.
\newblock A distributional perspective on reinforcement learning.
\newblock {\em arXiv preprint arXiv:1707.06887}.

\bibitem[\protect\astroncite{Chaslot et~al.}{2008}]{chaslot2008monte}
Chaslot, G., S.~Bakkes, I.~Szita, and P.~Spronck\leavevmode\nopagebreak\newline
  2008.
\newblock Monte-{Carlo} {Tree} {Search}: {A} {New} {Framework} for {Game} {AI}.

\bibitem[\protect\astroncite{Chuchro and Gupta}{2017}]{chuchrogame}
Chuchro, R. and D.~Gupta\leavevmode\nopagebreak\newline 2017.
\newblock Game {Playing} with {Deep} {Q}-{Learning} using {OpenAI} {Gym}.
\newblock P.~~6.

\bibitem[\protect\astroncite{Foley}{2017}]{foley2017model}
Foley, D.~J.\leavevmode\nopagebreak\newline 2017.
\newblock {\em Model-Based Reinforcement Learning in Atari 2600 Games}.
\newblock PhD thesis.

\bibitem[\protect\astroncite{Fortunato et~al.}{2017}]{fortunato_noisy_2017}
Fortunato, M., M.~G. Azar, B.~Piot, J.~Menick, I.~Osband, A.~Graves, V.~Mnih,
  R.~Munos, D.~Hassabis, O.~Pietquin, C.~Blundell, and
  S.~Legg\leavevmode\nopagebreak\newline 2017.
\newblock Noisy {Networks} for {Exploration}.
\newblock {\em arXiv:1706.10295 [cs, stat]}.
\newblock arXiv: 1706.10295.

\bibitem[\protect\astroncite{Fu and Hsu}{2016}]{fu2016model}
Fu, J. and I.~Hsu\leavevmode\nopagebreak\newline 2016.
\newblock Model-based reinforcement learning for playing atari games.
\newblock Technical report, Technical Report, Stanford University.

\bibitem[\protect\astroncite{Guo et~al.}{2014}]{guo2014deep}
Guo, X., S.~Singh, H.~Lee, R.~L. Lewis, and
  X.~Wang\leavevmode\nopagebreak\newline 2014.
\newblock Deep learning for real-time atari game play using offline monte-carlo
  tree search planning.
\newblock In {\em Advances in neural information processing systems}, Pp.~
  3338--3346.

\bibitem[\protect\astroncite{Hernández et~al.}{2017a}]{fractalai_repo}
Hernández, S., G.~Durán, and J.~M. Amigó\leavevmode\nopagebreak\newline
  2017a.
\newblock Fractal{AI}.
\newblock \url{https://github.com/FragileTheory/FractalAI}.

\bibitem[\protect\astroncite{Hernández et~al.}{2017b}]{hernandez2017general}
Hernández, S., G.~Durán, and J.~M. Amigó\leavevmode\nopagebreak\newline
  2017b.
\newblock General {Algorithmic} {Search}.
\newblock {\em arXiv:1705.08691 [math]}.
\newblock arXiv: 1705.08691.

\bibitem[\protect\astroncite{Hernández et~al.}{2018}]{cerezo2018fractal}
Hernández, S., G.~Durán, and J.~M. Amigó\leavevmode\nopagebreak\newline
  2018.
\newblock Fractal {AI}: {A} fragile theory of intelligence.
\newblock {\em arXiv:1803.05049 [cs]}.
\newblock arXiv: 1803.05049.

\bibitem[\protect\astroncite{Hester et~al.}{2017}]{hester2017deep}
Hester, T., M.~Vecerik, O.~Pietquin, M.~Lanctot, T.~Schaul, B.~Piot, D.~Horgan,
  J.~Quan, A.~Sendonaris, G.~Dulac-Arnold,
  et~al.\leavevmode\nopagebreak\newline 2017.
\newblock Deep q-learning from demonstrations.
\newblock {\em arXiv preprint arXiv:1704.03732}.

\bibitem[\protect\astroncite{Isasi et~al.}{2014}]{isasi2014schemata}
Isasi, P., M.~Drugan, and B.~Manderick\leavevmode\nopagebreak\newline 2014.
\newblock Schemata monte carlo network optimization.
\newblock In {\em PPSN 2014 Workshop: In Search of Synergies between
  Reinforcement Learning and Evolutionary Computation}.

\bibitem[\protect\astroncite{Mnih et~al.}{2016}]{mnih_asynchronous_2016}
Mnih, V., A.~P. Badia, M.~Mirza, A.~Graves, T.~P. Lillicrap, T.~Harley,
  D.~Silver, and K.~Kavukcuoglu\leavevmode\nopagebreak\newline 2016.
\newblock Asynchronous {Methods} for {Deep} {Reinforcement} {Learning}.
\newblock {\em arXiv:1602.01783 [cs]}.
\newblock arXiv: 1602.01783.

\bibitem[\protect\astroncite{Mnih et~al.}{2013}]{mnih2013playing}
Mnih, V., K.~Kavukcuoglu, D.~Silver, A.~Graves, I.~Antonoglou, D.~Wierstra, and
  M.~Riedmiller\leavevmode\nopagebreak\newline 2013.
\newblock Playing atari with deep reinforcement learning.
\newblock {\em arXiv preprint arXiv:1312.5602}.

\bibitem[\protect\astroncite{Mnih et~al.}{2015}]{mnih_human-level_2015}
Mnih, V., K.~Kavukcuoglu, D.~Silver, A.~A. Rusu, J.~Veness, M.~G. Bellemare,
  A.~Graves, M.~Riedmiller, A.~K. Fidjeland, G.~Ostrovski, S.~Petersen,
  C.~Beattie, A.~Sadik, I.~Antonoglou, H.~King, D.~Kumaran, D.~Wierstra,
  S.~Legg, and D.~Hassabis\leavevmode\nopagebreak\newline 2015.
\newblock Human-level control through deep reinforcement learning.
\newblock {\em Nature}, 518(7540):529--533.

\bibitem[\protect\astroncite{Plappert et~al.}{2017}]{plappert2017parameter}
Plappert, M., R.~Houthooft, P.~Dhariwal, S.~Sidor, R.~Y. Chen, X.~Chen,
  T.~Asfour, P.~Abbeel, and M.~Andrychowicz\leavevmode\nopagebreak\newline
  2017.
\newblock Parameter space noise for exploration.
\newblock {\em arXiv preprint arXiv:1706.01905}.

\bibitem[\protect\astroncite{Salimans et~al.}{2017}]{salimans2017evolution}
Salimans, T., J.~Ho, X.~Chen, S.~Sidor, and
  I.~Sutskever\leavevmode\nopagebreak\newline 2017.
\newblock Evolution strategies as a scalable alternative to reinforcement
  learning.
\newblock {\em arXiv preprint arXiv:1703.03864}.

\bibitem[\protect\astroncite{Segler et~al.}{2017}]{segler2017towards}
Segler, M., M.~Preu{\ss}, and M.~P. Waller\leavevmode\nopagebreak\newline 2017.
\newblock Towards" alphachem": Chemical synthesis planning with tree search and
  deep neural network policies.
\newblock {\em arXiv preprint arXiv:1702.00020}.

\bibitem[\protect\astroncite{Silver et~al.}{2016}]{silver2016mastering}
Silver, D., A.~Huang, C.~J. Maddison, A.~Guez, L.~Sifre, G.~Van Den~Driessche,
  J.~Schrittwieser, I.~Antonoglou, V.~Panneershelvam, M.~Lanctot,
  et~al.\leavevmode\nopagebreak\newline 2016.
\newblock Mastering the game of go with deep neural networks and tree search.
\newblock {\em nature}, 529(7587):484--489.

\bibitem[\protect\astroncite{Wissner-Gross and Freer}{2013}]{wissner2013causal}
Wissner-Gross, A.~D. and C.~E. Freer\leavevmode\nopagebreak\newline 2013.
\newblock Causal entropic forces.
\newblock {\em Physical review letters}, 110(16):168702.

\end{thebibliography}

\end{document}